\pdfoutput=1
\documentclass[letterpaper,journal]{IEEEtran}
\usepackage[T1]{fontenc}
\usepackage[utf8]{inputenc}
\usepackage{amsmath,amsfonts,amssymb,booktabs,graphicx,array,xcolor,url,multirow}
\IfFileExists{cite.sty}{\usepackage{cite}}{}
\usepackage[hidelinks,unicode]{hyperref}
\graphicspath{{figures/}}
\renewcommand{\arraystretch}{1.02}
\begin{document}
\raggedbottom
\title{ReliaGuard: Reconstruction-Aware Watermarking for Source Attribution and Blind Edit Localization}

\author{Hailong~Wang, Wuzheng~Tan, Zixin~Zhang, Xujun~Fang, and Fengjuan~Gong
\thanks{This work was supported by the National Natural Science
Foundation of China under Grant 62272199.
(Corresponding author: Wuzheng Tan.)}%
\thanks{Hailong Wang, Wuzheng Tan, Zixin Zhang, Xujun Fang,
and Fengjuan Gong are with the College of Cyber Security,
Jinan University, Guangzhou 510632, China}%
}

\markboth{IEEE Transactions on Multimedia}%
{Wang \MakeLowercase{\textit{et al.}}: ReliaGuard: Reconstruction-Aware Watermarking for Source Attribution and Blind Edit Localization}

\markboth{}{}

\maketitle

\begin{abstract}
Proactive image watermarking plays a vital role in safeguarding visual content against unauthorized generative editing by jointly enabling source attribution and tamper localization. However, latent diffusion editors frequently reconstruct unedited background regions outside the target mask, weakening spatial watermark evidence even when the source identifier remains recoverable. This creates a pervasive background reconstruction ambiguity in blind localization, where attenuated responses can stem from either intended edits or incidental background regeneration. To resolve this ambiguity, we propose ReliaGuard, a reconstruction-aware watermarking framework that coordinates reconstruction-stable evidence writing with coarse-to-fine blind reading. On the writing side, ReliaGuard aligns keyed spatial responses across both the protected image and its autoencoder reconstruction while preserving a frozen source-attribution channel under a quantization-aware distortion bound. On the reading side, a unified blind reader first refines a base localization map using image context and background priors, and then fuses native-resolution detail with multiscale representations under paired propagation training, requiring no reference image at inference. Extensive experiments across three image domains show that ReliaGuard achieves 85.61\% IoU on complete editing outputs and 94.90\% on locally spliced outputs at low background false-positive rates, outperforming the strongest baseline by 45.02 percentage points on complete outputs while achieving 100\% decoded identifier-bit accuracy. Across seven image-processing families, it maintains 81.01\% and 86.49\% localization IoU alongside strong identifier recovery, demonstrating a practical solution for joint attribution and blind edit localization under generative editing.
\end{abstract}

\begin{IEEEkeywords}
Generative image editing, image reconstruction, image watermarking, source attribution, tamper localization.
\end{IEEEkeywords}

\section{Introduction}

\IEEEPARstart{G}{enerative} image editors can replace objects, alter object appearance, and synthesize missing content while maintaining visual coherence~\cite{ldm}. Because edited outputs often remain visually close to their original scenes even after substantial semantic changes, practical content verification requires two complementary capabilities: robust watermarking to embed recoverable source information~\cite{hidden,stegastamp,vine}, and proactive localization to embed spatial signals whose alterations reveal where editing has occurred~\cite{editguard,wam,omniguard}.

Satisfying both requirements from a single received image without a reference places competing demands on the embedded signal. While the source identifier must remain globally recoverable after image processing, spatial localization must distinguish intended edits from incidental changes elsewhere in the image. This separation becomes particularly difficult in latent diffusion editing, where autoencoder encoding and decoding can also reconstruct background regions outside the requested mask~\cite{ldm,apt}. Following Anchor-aligned Perturbations for Tamper Localization (APT)~\cite{apt}, we distinguish full-output regeneration (FR), which retains the complete editor output, from local splicing (SP), which composites the same generated foreground onto the protected background. Although both outputs share identical foreground edits, their background evidence differs: a reader tuned for SP edits often fires across reconstructed FR backgrounds, whereas raising the decision threshold to suppress those false positives simultaneously erodes true edit recall.

We refer to this difficulty in separating target edits from reconstruction-induced background drift as \emph{background reconstruction ambiguity}. Although recent unified protection frameworks combine message recovery with spatial localization~\cite{editguard,omniguard,stableguard,genptw}, and APT explicitly strengthens feature alignment before full-output regeneration~\cite{apt}, preserving an embedded signal and interpreting its attenuation remain tightly coupled problems. Strengthening evidence against reconstruction makes intact backgrounds easier to preserve, while contextual reading helps separate localized edits from residual background variation. Spatial resolution is equally critical, because for small or irregular edits with high perimeter-to-area ratios, even modest boundary displacement causes substantial overlap loss.

ReliaGuard addresses these challenges through two complementary components: reconstruction-stable evidence writing (REW) and evidence-guided blind reading (EBR). On the writing side, REW embeds keyed spatial evidence into an image already carrying a source identifier from a frozen VINE-R encoder~\cite{vine}, jointly aligning raw and neighborhood-pooled responses across both the protected image and its variational autoencoder (VAE) reconstruction. By pairing a source-preservation objective with a quantization-aware distortion projection, REW allows spatial evidence and source attribution to share the same pixel budget without retraining the source encoder.

On the reading side, EBR interprets the received evidence through a unified coarse-to-fine hierarchy that first forms a base localization map from keyed responses and dual-scale image features, then refines that map using image context and an estimated background prior, and finally integrates native-resolution detail with coarser representations to recover accurate edit boundaries. Trained with paired supervision on unpropagated and propagated observations, EBR handles both reconstructed backgrounds and subsequent image processing, while a separate frozen decoder recovers the source identifier from the same received image. Neither branch requires the original image or a stored protected reference at inference.

We evaluate ReliaGuard across three image domains, four edit-area levels, and three mask-shape families, using paired FR and SP outputs to isolate the effect of background reconstruction. At thresholds calibrated to a nominal 1\% FR background false-positive rate (FPR), ReliaGuard achieves intersection-over-union (IoU) scores of 85.61\% on FR and 94.90\% on SP alongside 100\% decoded identifier-bit accuracy in both conditions. Additional experiments evaluate robustness across seven image-processing families and zero-shot transfer to an alternative editing protocol, while matched architectural ablations distinguish the contribution of each complete readout stage from that of its internal operators.

Our main contributions are summarized as follows:
\begin{itemize}
    \item We present ReliaGuard, a reconstruction-aware watermarking framework for source attribution and blind edit localization whose writing and reading components address background reconstruction ambiguity while retaining a frozen source-attribution channel and requiring no reference image at inference.

    \item We formulate dual-view evidence writing that jointly aligns raw and neighborhood-pooled keyed responses in protected and VAE-reconstructed images under source-preservation constraints, paired with a quantization-aware projection that enforces a per-image peak signal-to-noise ratio (PSNR) bound of 32~dB for the released eight-bit protected image.

    \item We develop a unified coarse-to-fine blind reader combining contextual residual refinement and native-guided multiscale decoding under paired propagation training, and validate its localization, source recovery, processing robustness, and cross-protocol transfer across three domains, with matched ablations isolating the contribution of each readout stage.
\end{itemize}
\section{Related Work}

\subsection{Robust Image Watermarking}

Learning-based watermarking typically trains an encoder--decoder pair separated by differentiable distortion layers. HiDDeN established this framework for robust message recovery using differentiable approximations to Joint Photographic Experts Group (JPEG) compression~\cite{hidden}, while StegaStamp extended robust decoding to physical printing and photographic recapture~\cite{stegastamp}. Subsequent studies further improved the fidelity--robustness trade-off: Mini-Batch of Real and Simulated JPEG compression (MBRS) mixes real JPEG, simulated JPEG, and distortion-free layers to strengthen compression tolerance~\cite{mbrs}, whereas TrustMark combines spatio-spectral supervision with residual resolution scaling for high-fidelity embedding at arbitrary resolutions~\cite{trustmark}.

A second line of work embeds watermarks in learned or generative representations. Feature-space watermarking optimizes image perturbations through a fixed self-supervised encoder~\cite{sslwm,apt}. In diffusion models, Stable Signature fine-tunes the VAE decoder~\cite{signature}, whereas Tree-Ring and Gaussian Shading embed watermark signals in the initial sampling noise~\cite{treering,gaussian}, and VINE combines generative priors with surrogate distortions to survive image editing~\cite{vine}. Because these methods focus on detection and message recovery rather than spatial edit boundaries, ReliaGuard retains the released VINE-R source-recovery branch and adds reconstruction-aware spatial evidence writing under an explicit source-preservation constraint.

\subsection{Proactive Localization and Unified Protection}

Proactive localization embeds spatial signals prior to distribution so that subsequent manipulations can be detected from signal degradation. EditGuard pairs localization-oriented watermarking with copyright message recovery~\cite{editguard}, and OmniGuard extends this framework with blind, degradation-aware tamper extraction and lightweight generative-editing simulation~\cite{omniguard}. Taking a segmentation perspective, the Watermark Anything Model (WAM) predicts watermarked regions directly while decoding region-associated messages~\cite{wam}. Recent approaches also couple watermarking with generative representations: StableGuard jointly optimizes an adapted VAE decoder and a mixture-of-experts forensic network~\cite{stableguard}, while GenPTW integrates latent-space embedding with provenance tracing and tamper localization in both generation-time and post-generation settings~\cite{genptw}.

When generative editors reconstruct the entire output, background changes outside the target region severely challenge localization. APT addresses fully regenerated outputs by optimizing latent perturbations aligned with an anchor direction, and APT$^*$ appends a shallow decoder to refine the resulting alignment map~\cite{apt}. ReliaGuard similarly adopts feature-space evidence, but jointly optimizes raw and neighborhood-pooled keyed responses across protected and VAE-reconstructed views while preserving a separate source identifier, and interprets the received responses through contextual and multiscale blind reading.

\subsection{Forensic Features and Spatial Decoding}

Passive image forensics localizes manipulations solely from intrinsic traces in the received image. Noiseprint extracts camera-model-related residuals by suppressing scene content~\cite{noiseprint}, CAT-Net fuses red, green, and blue (RGB) features with discrete cosine transform (DCT) cues to expose compression inconsistencies~\cite{catnet}, and MVSS-Net combines noise and boundary cues through multi-view, multiscale supervision~\cite{mvss}. More recently, TruFor integrates visual and forensic representations with pixelwise reliability estimation~\cite{trufor}, while IML-ViT adapts vision transformers to high-resolution, multiscale manipulation localization~\cite{imlvit}. Unlike passive detectors, proactive frameworks can combine deliberately embedded spatial evidence with visual context.

Dense-prediction architectures supply complementary mechanisms for balancing global context with fine spatial detail, including self-supervised dense features~\cite{dinov3}, encoder--decoder skip connections~\cite{unet}, non-local context aggregation~\cite{nonlocal}, and content-aware feature reassembly~\cite{carafe}. Our unified blind reader adapts these mechanisms to interpret keyed watermark evidence in two successive stages: contextual residual refinement to distinguish local edits from reconstructed backgrounds, followed by native-guided multiscale decoding to preserve sharp spatial contours.
\section{Method}
\label{sec:method}

\begin{figure*}[tp]
\centering
\includegraphics[
    width=\textwidth,
    height=.57\textheight,
    keepaspectratio
]{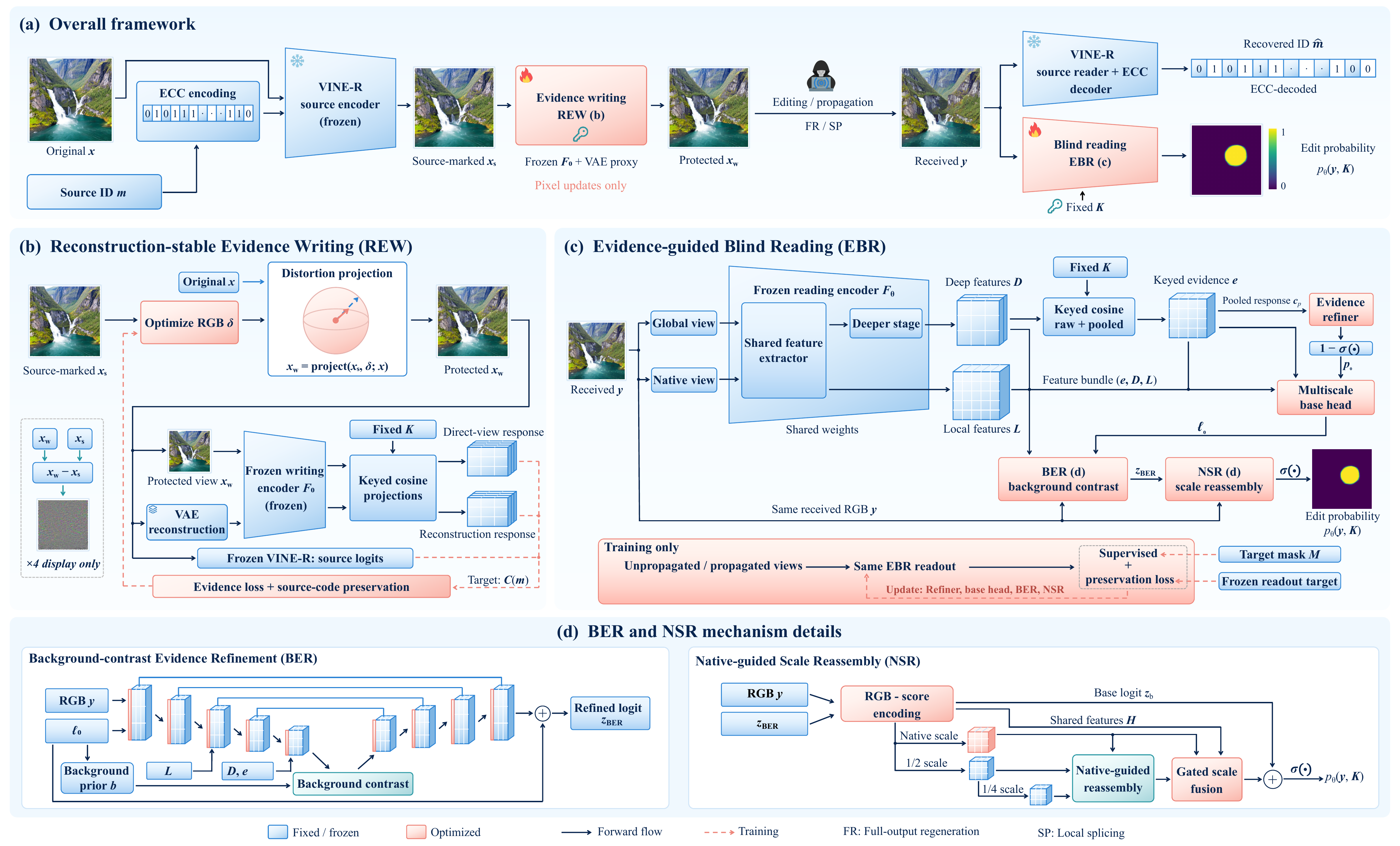}
\caption{
Overview of the proposed ReliaGuard framework.
(a) Source-message encoding and REW produce the protected image $x_w$; a frozen source decoder and EBR recover the identifier $\widehat{m}$ and edit probability $p_\Theta(y,K)$ without a reference image.
(b) REW optimizes keyed responses in protected and VAE-reconstructed views under source-preservation and eight-bit quantization-aware distortion constraints.
(c) EBR extracts dual-scale features and keyed evidence from $y$ and is trained on paired unpropagated and propagated observations.
(d) Internal stages of EBR: contextual residual refinement (BER) and native-guided multiscale readout (NSR).
Solid arrows denote forward inference; dashed arrows denote optimization or training supervision.
Blue and coral blocks indicate fixed and optimized components, respectively. Error-correcting coding (ECC) is used for the source message.
}
\label{fig:pipeline}
\end{figure*}

As illustrated in Fig.~\ref{fig:pipeline}, ReliaGuard addresses background reconstruction ambiguity through two complementary components: reconstruction-stable evidence writing (REW) and evidence-guided blind reading (EBR). On the writing side, REW aligns keyed spatial responses in both the protected image and its VAE reconstruction while preserving a frozen source-attribution channel under an eight-bit quantization-aware distortion bound. On the reading side, a separate frozen decoder recovers the source identifier, while EBR localizes edits in a unified coarse-to-fine hierarchy consisting of dual-scale evidence extraction, contextual residual refinement (BER), and native-guided multiscale readout (NSR).

\subsection{Problem Formulation}

Let $x\in[0,1]^{3\times H\times W}$ be an original image, $m\in\{0,1\}^{16}$ its source identifier, and $K$ a fixed unit-norm evidence key. A frozen VINE-R encoder~\cite{vine} first embeds the error-correcting-coded identifier $\mathcal C(m)$ (96 bits, padded with four fixed zeros to match the 100-bit interface) to obtain $x_s$. REW then adds spatial evidence under a source-preservation constraint to produce the protected image $x_w$.

Given an editing prompt $t$, a binary target mask $M\in\{0,1\}^{H\times W}$, and a random seed $\xi$, we construct
\begin{align}
 y_F&=\mathcal E(x_w,M,t;\xi),
 \label{eq:fr}\\
 y_S&=M\odot y_F+(1-M)\odot x_w,
 \label{eq:sp}
\end{align}
where $\mathcal E$ denotes the generative editor and $\odot$ denotes elementwise multiplication broadcast across color channels. FR retains the complete editor output $y_F$, whereas SP composites its generated foreground onto the protected background~\cite{apt}. Both outputs share the same foreground within $M$ and the same localization target $M$, isolating the effect of background reconstruction.

An optional propagation operation $\mathcal A$ produces the received image $y$. The source decoder recovers $\widehat m$ from the 96 coded-bit logits and outputs an acceptance confidence. The localization branch maps $(y,K)$ to a pixelwise edit probability $p_\Theta(y,K)$ with binary prediction
\begin{equation}
 \widehat M=\mathbf1[p_\Theta(y,K)\ge\tau].
\end{equation}
Each reader uses a single threshold $\tau$ calibrated on a separate calibration set and kept fixed across output types and propagation conditions. Neither branch requires the original or protected reference image at inference. During training and evaluation, geometric transformations act jointly on the image, target mask, and valid support.

\subsection{Reconstruction-Stable Dual-View Evidence Writing (REW)}

Weak background evidence makes a local edit difficult to separate from global reconstruction after editing. REW addresses this issue at writing time by jointly optimizing keyed alignment in both $x_w$ and its deterministic VAE reconstruction $R(x_w)$. Following reconstruction-based degradation modeling~\cite{omniguard}, we decode the posterior mean of the Stable Diffusion 2 (SD2) inpainting VAE:
\begin{equation}
 R(x_w)=\operatorname{clip}\!\left(
 \frac{D_{\mathrm{VAE}}(\mu_{\mathrm{VAE}}(2x_w-1))+1}{2},0,1
 \right),
 \label{eq:vae_reconstruction}
\end{equation}
where $\mu_{\mathrm{VAE}}$ and $D_{\mathrm{VAE}}$ denote the VAE encoder's posterior mean and decoder, respectively. Decoding the posterior mean keeps reconstruction deterministic during writing.

A frozen writing encoder $F_0$ maps each resized view to spatial features $f$. Following feature-space watermarking~\cite{sslwm,apt}, we measure alignment with $K$ before and after neighborhood pooling:
\begin{align}
 c_r(u)&=\langle\nu(f(u)),K\rangle,\\
 c_p(u)&=
 \left\langle
 \nu\!\left(\operatorname{Avg}_{3\times3}f(u)\right),K
 \right\rangle,
\end{align}
where $\nu$ denotes channel-wise $\ell_2$ normalization and $\operatorname{Avg}_{3\times3}$ is stride-one zero-padded average pooling. Because pooling precedes normalization, both local feature directions and magnitudes affect $c_p$.

Concatenating the raw and pooled response maps into $\{c_i\}_{i=1}^{N_e}$ and setting $a_i=\operatorname{softplus}[-\kappa(c_i-\mu)]$, we define the evidence loss as
\begin{equation}
 \mathcal L_e=
 \frac{1}{N_e}\sum_{i=1}^{N_e}a_i+
 \frac{1}{|\mathcal H|}\sum_{i\in\mathcal H}a_i,
\end{equation}
where $\mathcal H$ indexes the top $\max(1,\lfloor0.1N_e\rfloor)$ largest losses, placing greater weight on weakly aligned regions. The writing objective is
\begin{equation}
\begin{aligned}
 \mathcal L_w={}&
 \tfrac12\left[
 \mathcal L_e(x_w)+\mathcal L_e(R(x_w))
 \right]+\mathcal L_{\mathrm{src}}(x_w),
\end{aligned}
\label{eq:write}
\end{equation}
where $\mathcal L_{\mathrm{src}}$ is the binary cross-entropy between the frozen source decoder's 96 coded-bit logits and $\mathcal C(m)$ evaluated on $x_w$.

REW optimizes a zero-initialized RGB increment $\delta$ while keeping the source codec, $F_0$, and VAE fixed. Let $Q_8$ denote nearest-level eight-bit quantization and $\|\cdot\|_{\mathrm{RMS}}$ the root mean square over all color values. At each update, we apply a quantization-aware distortion projection:
\begin{align}
 u&=\operatorname{clip}\bigl(
 x_s+\operatorname{clip}(\delta,-16/255,16/255),0,1
 \bigr),\\
 r&=10^{-32/20}-1/(2\cdot255),\\
 \alpha&=\min\{1,r/(\|u-x\|_{\mathrm{RMS}}+\epsilon)\},\\
 x_w&=Q_8\bigl(x+\alpha(u-x)\bigr).
\end{align}
By reserving the maximum quantization error $1/(2\cdot255)$, the projection ensures $\|x_w-x\|_{\mathrm{RMS}}\le 10^{-32/20}$, guaranteeing a PSNR of at least 32~dB for the released eight-bit image. Gradients through $Q_8$ are estimated using a straight-through estimator.

\subsection{Evidence-Guided Coarse-to-Fine Blind Reading (EBR)}

Preserving the embedded signal is only part of the localization task, because residual attenuation across textured backgrounds can still overlap with weak foreground responses. EBR therefore formulates blind localization as a single coarse-to-fine hierarchy comprising three successive stages: dual-scale feature and evidence extraction, contextual residual refinement, and native-guided multiscale readout.

\subsubsection{Dual-Scale Feature and Keyed Evidence Extraction}
A shared DINOv3 ConvNeXt reading encoder $F_\theta$~\cite{dinov3,convnext} (with weights separate from $F_0$) processes a reduced global view and the native-resolution received image. Both paths share the input stem and first stage to yield local features $L$, while the global path continues through the second stage to yield deep features $D$. Raw and neighborhood-pooled projections of $D$ onto $K$ form the two-channel evidence map $e$. An evidence refiner $R_e$ converts the pooled response into a preliminary edit probability
\begin{equation}
 p_0=1-\sigma(R_e(c_p)),
\end{equation}
and a multiscale base head $H_b$ combines $(e,p_0,D,L)$ to produce base logits $\widetilde\ell_0$, stabilized as
\begin{equation}
 \ell_0=\operatorname{logit}\!\left(
 \operatorname{clip}(
 \sigma(\widetilde\ell_0),10^{-6},1-10^{-6}
 )\right).
\end{equation}

\subsubsection{Contextual Residual Refinement (BER Stage)}
A uniformly weak response does not carry the same meaning as a localized drop relative to its surroundings. The BER stage refines $\ell_0$ using image context and an estimate of the intact background. A skip-connected encoder--decoder~\cite{unet} takes $[2y-1,\ell_0/8]$ as input, injecting $L$ at an intermediate stage and $(D,e)$ at the bottleneck. The base prediction defines soft background weights on the bottleneck grid:
\begin{equation}
 b=\max\!\left\{
 \operatorname{Pool}\!\left(
 1-\sigma(\operatorname{sg}(\ell_0))
 \right),10^{-3}
 \right\},
\end{equation}
where $\operatorname{Pool}$ denotes adaptive average pooling and $\operatorname{sg}$ stops gradients. Let $h_i$ be a bottleneck feature with channel-normalized projections $(q_i,k_i,v_i)$ of dimension $d$. Background-weighted non-local aggregation~\cite{nonlocal} updates each location via
\begin{align}
 A_{ij}&=
 \frac{\exp(q_i^\top k_j/\sqrt d+\log b_j)}
 {\sum_l\exp(q_i^\top k_l/\sqrt d+\log b_l)},\\
 h'_i&=
 h_i+P\!\left(\sum_j A_{ij}v_j-v_i\right),
\end{align}
so that the update depends on the contrast between local features and the estimated background context. The decoder outputs a residual logit correction:
\begin{equation}
 z_{\mathrm{BER}}=
 \ell_0+\Delta_{\mathrm{BER}}(y,\ell_0,e,D,L).
 \label{eq:ber}
\end{equation}
Only the background-prior branch is detached; gradients remain active through the encoder input and the identity shortcut.

\subsubsection{Native-Guided Multiscale Readout (NSR Stage)}
Contextual refinement identifies edited regions, whereas coarse features alone can blur small targets and irregular contours. The NSR stage therefore retains a native-resolution path alongside lower-resolution context. An RGB-and-score encoder constructs a shared representation $H$ from $2y-1$, $\operatorname{clip}(z_{\mathrm{BER}}/8,-2,2)$, and four normalized spatial-derivative channels of $z_{\mathrm{BER}}$:
\begin{equation}
 d_x=\partial_x(2z_{\mathrm{BER}}),\qquad
 d_y=\partial_y(2z_{\mathrm{BER}}),
\end{equation}
\begin{equation}
 s=\sqrt{d_x^2+d_y^2+10^{-6}},
 \qquad n_x=d_x/s,\qquad n_y=d_y/s,
\end{equation}
\begin{equation}
 \left[
 n_x,\,
 n_y,\,
 \tanh(s/2),\,
 \operatorname{clip}\bigl(
 4(\partial_xn_x+\partial_yn_y),-1,1
 \bigr)
 \right].
\end{equation}
Three encoding blocks extract native-, half-, and quarter-resolution features:
\begin{align}
 n&=P_N(H),\qquad
 h=P_H(\operatorname{Avg}_2H),\\
 q&=P_Q(\operatorname{Avg}_2h).
\end{align}
After channel projection, each lower-resolution feature map $l\in\{h,q\}$ is upsampled under native-feature guidance~\cite{carafe}:
\begin{equation}
\begin{aligned}
 \mathcal R(l,n)={}&U_{\mathrm{bilinear}}(l)\\
 &+\sum_{j=1}^{9}
 \left[
 \operatorname{softmax}_j(\psi(l,n))-\frac19
 \right]U_{\mathrm{nearest}}(l_j),
\end{aligned}
\label{eq:reassembly}
\end{equation}
producing $\widetilde h$ and $\widetilde q$, where $l_j$ indexes the $3\times3$ neighborhood with replicate padding and the final layer of $\psi$ is zero-initialized. Pixelwise scale weights $a=\operatorname{softmax}\bigl(G(H)+G_\Delta([n,\widetilde h,\widetilde q])\bigr)$ fuse the three scales, while a parallel score branch computes a gated correction on nonsaturated logits:
\begin{equation}
\begin{aligned}
 z_b={}&
 z_{\mathrm{BER}}+
 2\eta(z_{\mathrm{BER}})\tanh(O(H))+R_b(H),
\end{aligned}
\label{eq:nsr_base}
\end{equation}
with $\eta(z)=\operatorname{clip}(1-|z|/8,0,1)$. The final edit probability is
\begin{equation}
\begin{aligned}
 z&=z_b+C(a_1n+a_2\widetilde h+a_3\widetilde q),\\
 p_\Theta(y,K)&=\sigma(z).
\end{aligned}
\label{eq:nsr}
\end{equation}

\subsection{Paired Propagation Training}
\label{sec:training}

A reader trained only on unprocessed images can rely on fragile cues that degrade during image distribution. EBR is therefore trained in two phases: first, $F_\theta$ and the readout stages are adapted on unpropagated FR/SP pairs and unedited protected images; next, $F_\theta$ is frozen and $R_e$, $H_b$, BER, and NSR are jointly optimized on paired unpropagated and propagated observations.

Each source contributes two FR, two SP, and one unedited (\emph{clean}) observation per epoch, each paired with a propagated counterpart. Geometric transformations act jointly on the image, mask, and valid support, and crops with no remaining foreground serve as negative observations. For logits $z$, probabilities $p=\sigma(z)$, target $g$, and valid support $v$, let $b_i=\operatorname{softplus}(z_i)-g_i z_i$. The region-and-boundary loss combines cross-entropy, soft Dice~\cite{vnet}, and inner/outer boundary cross-entropy:
\begin{equation}
\begin{aligned}
 \mathcal L_b={}&
 \frac{\sum_i v_i b_i}{\max(1,\sum_i v_i)}\\
 &+0.5\left(
 1-\frac{2\sum_i v_i p_i g_i+\epsilon}
 {\sum_i v_i(p_i+g_i)+\epsilon}
 \right)+0.1\mathcal L_\partial,
\end{aligned}
\end{equation}
where stride-one $9\times9$ max pooling $\mathcal P_9$ defines inner and outer boundary supports $I=gv\,\mathbf1[\mathcal P_9(1-g)>0]$ and $O=(1-g)v\,\mathbf1[\mathcal P_9(g)>0]$, and
\begin{equation}
 \mathcal L_\partial=
 \tfrac12\left(
 \frac{\sum_i I_i b_i}{\max(1,\sum_i I_i)}+
 \frac{\sum_i O_i b_i}{\max(1,\sum_i O_i)}
 \right).
\end{equation}
During paired training, edited observations use $1.25\mathcal L_b$, while unedited images and empty-foreground crops use $0.5$ times the mean valid cross-entropy.

A frozen copy of the initialized reader supplies preservation targets $q_i$ on unpropagated views over high-confidence correct pixels
\begin{equation}
\begin{aligned}
 \mathcal I_c=\{i:\;&
 g_i=c,\ \max(q_i,1-q_i)\ge0.9,\ \mathbf1[q_i\ge0.5]=g_i\},
\end{aligned}
\end{equation}
yielding the class-balanced Bernoulli KL preservation loss
\begin{equation}
 \mathcal L_{\mathrm{keep}}=
 \frac12\sum_{c\in\{0,1\}}
 \frac{
 \sum_{i\in\mathcal I_c}
 D_{\mathrm{KL}}\bigl(
 \operatorname{Bern}(q_i)\|
 \operatorname{Bern}(p_i)
 \bigr)
 }{\max(1,|\mathcal I_c|)}.
\end{equation}
The complete training objective is
\begin{equation}
 \mathcal L=
 (1-\lambda_p)\mathcal L_n+
 \lambda_p\mathcal L_p+
 0.05\mathcal L_{\mathrm{keep}},
 \label{eq:train}
\end{equation}
where $\mathcal L_n$ and $\mathcal L_p$ supervise unpropagated and propagated views. The propagation weight $\lambda_p$ stays at $0.25$ for epochs 1--5, increases linearly to $0.5$ over epochs 6--15, and remains at $0.5$ thereafter, paired with a 15-epoch augmentation-severity ramp covering JPEG compression, resizing, blur, noise, color adjustment, geometric transforms, and two-step chains. At inference, both the source decoder and EBR operate directly on $y$.
\section{Experiments}
\label{sec:experiments}

In this section, we evaluate whether reconstruction-aware writing and coarse-to-fine blind reading retain source attribution while localizing edits under reconstructed backgrounds. We first compare localization and message recovery across domains, then evaluate background-error calibration, processing robustness, protected-image quality, target geometry, and cross-protocol transfer. Ablation studies isolate the written evidence, the complete readout stages, and their internal operators.

\subsection{Experimental Setup}

\subsubsection{Datasets and Evaluation Protocol}
We use Common Objects in Context (COCO)~\cite{coco}, ImageNet~\cite{imagenet}, and generated images, with 606 sources for training, 72 for calibration, and 144 for ablation evaluation. The main comparison is conducted on an independent test set of 300 sources (100 per domain, totaling 900 editing cases). The masks cover three shape categories and four area ratios: 3\%, 8\%, 18\%, and 35\%. Editing uses SD2 inpainting at $512\times512$ with 30 denoising steps, strength 1.0, and guidance scale 7.5, producing paired FR and SP outputs.

We select six representative active protection methods as baselines: WAM~\cite{wam}, StableGuard-C~\cite{stableguard}, OmniGuard~\cite{omniguard}, EditGuard~\cite{editguard}, GenPTW~\cite{genptw}, and APT~\cite{apt}, together with APT's enhanced reader APT$^*$. Two passive manipulation-localization methods, TruFor~\cite{trufor} and IML-ViT~\cite{imlvit}, are also included. StableGuard-C uses its released clean-trained configuration; on unpropagated images, it attains 98.16\% SP IoU but drops to 16.73\% fixed-threshold FR IoU (0.71\% after calibration), showing that its FR degradation stems from VAE background reconstruction rather than post-processing. All baselines use released pretrained weights, and active methods retain their original writing procedures and payloads.

Localization is measured using IoU, F1, and background FPR, together with eight-pixel inner-band false-negative rate (FNR) and outer-band FPR. Source recovery is measured by bit accuracy (BA) and extraction rate (ER). Image fidelity is evaluated using PSNR, structural similarity index (SSIM)~\cite{ssim}, and learned perceptual image patch similarity (LPIPS)~\cite{lpips}. We report results at fixed thresholds (0.5, or 0.2 on EditGuard's raw residual) and at reader-specific thresholds calibrated to a nominal 1\% FR background FPR, held fixed across FR, SP, and processing conditions. Scores are averaged per source with equal weighting across the three domains. Following the broad distortion coverage used in watermark evaluation~\cite{waves}, robustness tests cover 18 settings across seven processing families (averaged within each family and then equally across families), with four stronger distortions evaluated separately. Paired 95\% confidence intervals use 5,000 within-domain source-group bootstrap resamples.

\subsubsection{Implementation Details}
REW runs 200 Adam updates per image with learning rate $0.002$, margin $\mu=0.1$, and steepness $\kappa=40$. The reading encoder $F_\theta$ takes a $256\times256$ global view and a $512\times512$ native view, producing $D\in\mathbb R^{192\times32\times32}$ and $L\in\mathbb R^{96\times128\times128}$. BER receives local features at $128\times128$ and deep/keyed features at the $32\times32$ bottleneck with query/key dimension $d=32$, and lower-resolution NSR paths use 24 channels after projection. The readout is trained for 100 epochs using Adam with decoupled weight decay (AdamW)~\cite{adamw} and an effective batch size of 24 observation pairs, with initial learning rates of $2\times10^{-5}$ for $R_e$ and $H_b$ and $5\times10^{-5}$ for BER and NSR.

\subsection{Comparison with Existing Methods}

\subsubsection{Localization and Source Recovery}
\begin{table*}[tp]
\centering
\caption{Localization and source-identifier recovery of active and passive methods across three image domains at fixed thresholds.}
\label{tab:main}
\label{tab:domains}
\fontsize{8}{10}\selectfont
\setlength{\tabcolsep}{3.5pt}
\renewcommand{\arraystretch}{1.12}

\begin{tabular*}{\linewidth}{@{\extracolsep{\fill}}l*{9}{c}@{}}
\toprule
\multirow[c]{2}{*}[-0.6ex]{Method} &
\multicolumn{3}{c}{COCO ($n=100$)} &
\multicolumn{3}{c}{Generated ($n=100$)} &
\multicolumn{3}{c}{ImageNet ($n=100$)} \\
\cmidrule(lr){2-4}
\cmidrule(lr){5-7}
\cmidrule(lr){8-10}
& IoU$\uparrow$ & F1$\uparrow$ & $\mathrm{BA}_{\mathrm{ID}}\uparrow$
& IoU$\uparrow$ & F1$\uparrow$ & $\mathrm{BA}_{\mathrm{ID}}\uparrow$
& IoU$\uparrow$ & F1$\uparrow$ & $\mathrm{BA}_{\mathrm{ID}}\uparrow$ \\
\midrule

\multicolumn{10}{c}{\textbf{Full-output regeneration (FR)}} \\
\midrule
\multicolumn{10}{c}{\textit{Passive localization}} \\
TruFor & 4.35 & 7.20 & -- & 4.89 & 8.11 & -- & 5.23 & 8.53 & -- \\
IML-ViT & 1.39 & 2.45 & -- & 1.43 & 2.56 & -- & 2.04 & 3.54 & -- \\

\addlinespace[2pt]
\multicolumn{10}{c}{\textit{Active protection}} \\
WAM & 25.10 & 37.08 & 98.58$^{\dagger}$ & 18.27 & 28.75 & 90.12$^{\dagger}$ & 27.24 & 39.71 & 96.70$^{\dagger}$ \\
StableGuard-C & 16.74 & 26.72 & 50.25 & 16.73 & 26.70 & 49.92 & 16.70 & 26.66 & 50.54 \\
OmniGuard & 0.41 & 0.79 & -- & 0.43 & 0.83 & -- & 0.58 & 1.12 & -- \\
EditGuard & 16.00 & 25.75 & 53.44 & 16.00 & 25.75 & 54.37 & 16.00 & 25.75 & 54.50 \\
GenPTW & 16.02 & 25.78 & \underline{98.60} & 16.01 & 25.76 & \underline{98.52} & 16.24 & 26.07 & \underline{99.25} \\
APT & 42.87 & 56.25 & -- & 42.97 & 56.73 & -- & 39.59 & 52.31 & -- \\
APT$^*$ & \underline{53.37} & \underline{65.18} & -- & \underline{47.89} & \underline{59.73} & -- & \underline{52.89} & \underline{65.00} & -- \\
ReliaGuard & \textbf{85.03} & \textbf{91.42} & \textbf{100.00} & \textbf{85.12} & \textbf{91.51} & \textbf{100.00} & \textbf{86.92} & \textbf{92.60} & \textbf{100.00} \\
\midrule

\multicolumn{10}{c}{\textbf{Local splicing (SP)}} \\
\midrule
\multicolumn{10}{c}{\textit{Passive localization}} \\
TruFor & 13.01 & 18.72 & -- & 7.76 & 12.37 & -- & 63.09 & 72.01 & -- \\
IML-ViT & 5.66 & 8.34 & -- & 5.04 & 8.04 & -- & 29.90 & 37.22 & -- \\

\addlinespace[2pt]
\multicolumn{10}{c}{\textit{Active protection}} \\
WAM & 77.10 & 83.11 & \textbf{100.00} & 77.22 & 83.78 & \textbf{100.00} & 78.46 & 84.66 & \textbf{100.00} \\
StableGuard-C & \textbf{98.24} & \textbf{99.08} & \textbf{100.00} & \textbf{98.35} & \textbf{99.16} & \textbf{100.00} & \textbf{97.90} & \textbf{98.77} & \textbf{100.00} \\
OmniGuard & 88.19 & 92.26 & -- & 88.35 & 92.39 & -- & 81.37 & 86.30 & -- \\
EditGuard & 80.72 & 87.88 & \textbf{100.00} & 82.96 & 89.74 & \textbf{100.00} & 81.67 & 88.26 & \textbf{100.00} \\
GenPTW & 90.73 & 92.86 & \underline{99.98} & 91.32 & 93.55 & \textbf{100.00} & 91.55 & 93.53 & \textbf{100.00} \\
APT & 41.20 & 53.24 & -- & 48.13 & 60.74 & -- & 41.07 & 52.71 & -- \\
APT$^*$ & 63.02 & 73.67 & -- & 59.16 & 70.23 & -- & 64.77 & 75.60 & -- \\
ReliaGuard & \underline{93.86} & \underline{96.60} & \textbf{100.00} & \underline{94.17} & \underline{96.79} & \textbf{100.00} & \underline{96.81} & \underline{98.26} & \textbf{100.00} \\
\bottomrule
\end{tabular*}

\par\smallskip
{\fontsize{7}{8.5}\selectfont\raggedright
Scores are percentages. Thresholds are 0.5, except for EditGuard's raw residual threshold of 0.2.
$\mathrm{BA}_{\mathrm{ID}}$ is evaluated on extracted messages.
$\dagger$ WAM's FR extraction rates are 98.33\%, 81.67\%, and 96.33\% for COCO, Generated, and ImageNet, respectively; the other reported message outputs have 100\% extraction.
-- denotes no comparable identifier result.
\par}
\end{table*}

Table~\ref{tab:main} compares localization and source-identifier recovery across the three image domains at fixed thresholds. ReliaGuard achieves the highest FR IoU in every domain, obtaining 85.03\% on COCO, 85.12\% on generated images, and 86.92\% on ImageNet, with a domain-averaged FR IoU of 85.69\% and F1 of 91.84\%. Under SP, these scores increase to 94.95\% and 97.22\%. StableGuard-C obtains higher SP overlap, whereas its FR overlap drops sharply once the reconstructed background is retained. ReliaGuard maintains high overlap in both conditions, with a smaller gap between FR and SP.

The passive baselines show a pronounced drop between SP and FR. On ImageNet, TruFor and IML-ViT obtain 63.09\% and 29.90\% SP IoU, respectively, but only 5.23\% and 2.04\% FR IoU, with similarly low FR scores on the other two domains. This behavior indicates that passive forensic cues are less effective at separating localized edits from a fully reconstructed background.

Decoded identifier-bit accuracy and ER both reach 100\% on unpropagated FR and SP outputs. By comparison, WAM's FR identifier-bit accuracy ranges from 90.12\% to 98.58\% across domains, with extraction rates of 81.67\%--98.33\%. The improvement in FR localization is therefore achieved alongside reliable identifier recovery.

\begin{figure*}[tp]
\centering
\includegraphics[width=.97\textwidth]{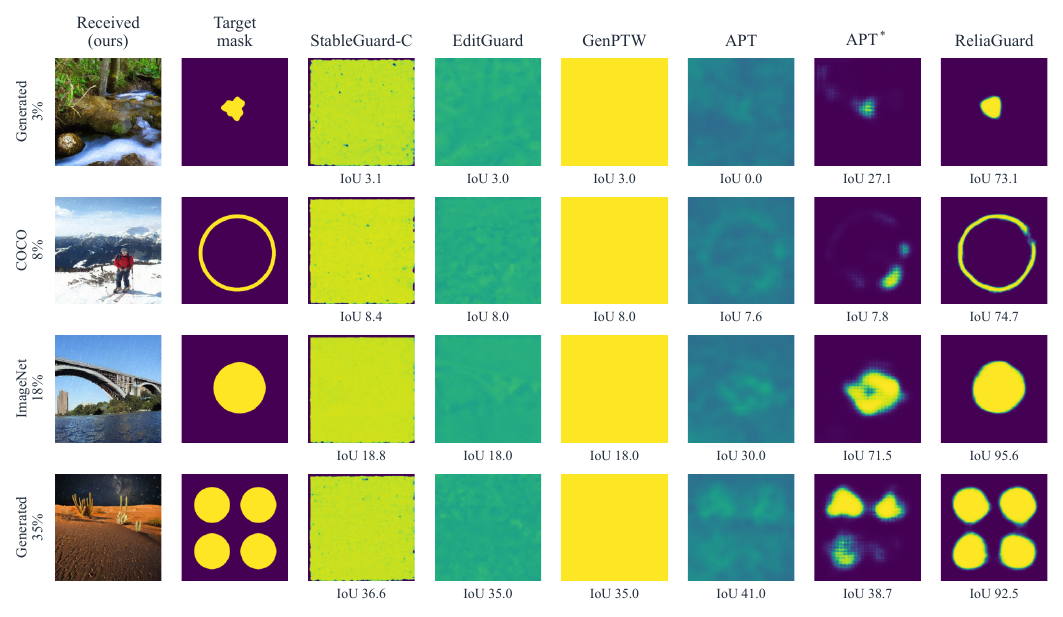}
\caption{FR localization examples for active methods across three image domains and four target areas. Each row uses the same source and target mask across methods; each method receives its own protected and edited image. The first column shows our received image. IoUs are evaluated at the fixed thresholds in Table~\ref{tab:main}.}
\label{fig:qual}
\end{figure*}

\begin{figure*}[tp]
\centering
\includegraphics[width=.97\textwidth]{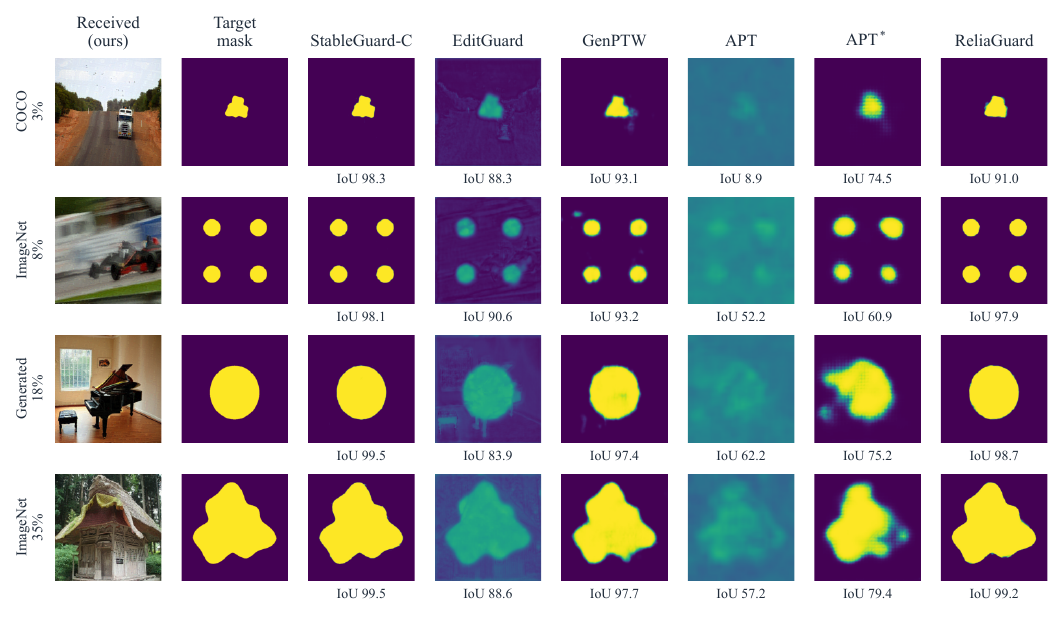}
\caption{SP localization examples for active methods across three image domains and four target areas. These cases differ from those in Fig.~\ref{fig:qual}; IoUs use the same fixed-threshold convention.}
\label{fig:qual_sp}
\end{figure*}

Qualitative comparisons in Figs.~\ref{fig:qual} and~\ref{fig:qual_sp} illustrate these differences. Several baselines respond across large portions of the reconstructed background in FR, whereas ReliaGuard concentrates its predictions around the target edit. On the small irregular waterfall example, ReliaGuard achieves 73.07\% IoU compared with 27.09\% for APT$^*$. On the SP piano and temple examples, ReliaGuard reaches 98.69\% and 99.25\%, respectively. Remaining errors appear mainly along thin boundaries and small disconnected regions.

\subsubsection{Localization at Low Background Error}
\begin{table*}[tp]
\centering
\caption{Localization at fixed thresholds and at thresholds calibrated to a nominal 1\% FR background FPR.}
\label{tab:thresholds}
\fontsize{8}{10}\selectfont
\setlength{\tabcolsep}{3.5pt}
\renewcommand{\arraystretch}{1.12}

\begin{tabular*}{\linewidth}{@{\extracolsep{\fill}}l*{6}{c}@{}}
\toprule
\multirow[c]{2}{*}[-0.6ex]{Method} &
\multicolumn{3}{c}{FR} &
\multicolumn{3}{c}{SP} \\
\cmidrule(lr){2-4}
\cmidrule(lr){5-7}
& IoU$\uparrow$ & F1$\uparrow$ & FPR$\downarrow$
& IoU$\uparrow$ & F1$\uparrow$ & FPR$\downarrow$ \\
\midrule

\multicolumn{7}{c}{\textit{Fixed decision thresholds}} \\
\midrule
WAM & 23.54 & 35.18 & 60.76 & 77.60 & 83.85 & 0.97 \\
StableGuard-C & 16.73 & 26.69 & 94.61 & \textbf{98.16} & \textbf{99.00} & 0.23 \\
OmniGuard & 0.47 & 0.91 & 0.31 & 85.97 & 90.32 & 0.36 \\
EditGuard & 16.00 & 25.75 & 100.00 & 81.78 & 88.63 & 4.93 \\
GenPTW & 16.09 & 25.87 & 99.46 & 91.20 & 93.31 & 0.30 \\
APT & 41.81 & 55.09 & 5.35 & 43.47 & 55.56 & 1.09 \\
APT$^*$ & \underline{51.39} & \underline{63.30} & 3.45 & 62.32 & 73.17 & 3.02 \\
ReliaGuard & \textbf{85.69} & \textbf{91.84} & 1.25 & \underline{94.95} & \underline{97.22} & 0.27 \\
\midrule

\multicolumn{7}{c}{\textit{FR-calibrated thresholds (nominal 1\% background FPR)}} \\
\midrule
WAM & 1.50 & 2.68 & 0.92 & 3.61 & 6.39 & 0.00 \\
StableGuard-C & 0.71 & 1.28 & 1.01 & \textbf{96.47} & \textbf{98.10} & 0.10 \\
OmniGuard & 1.29 & 2.42 & 1.07 & 86.18 & 90.77 & 0.50 \\
EditGuard & 0.97 & 1.83 & 0.90 & 0.62 & 1.15 & 0.00 \\
GenPTW & 0.57 & 1.11 & 1.06 & 0.00$^{\dagger}$ & 0.00$^{\dagger}$ & 0.00 \\
APT & 17.93 & 26.47 & 0.54 & 16.28 & 23.93 & 0.08 \\
APT$^*$ & \underline{40.59} & \underline{51.54} & 1.02 & 49.78 & 60.77 & 0.68 \\
ReliaGuard & \textbf{85.61} & \textbf{91.80} & 1.07 & \underline{94.90} & \underline{97.18} & 0.24 \\
\bottomrule
\end{tabular*}

\par\smallskip
{\fontsize{7}{8.5}\selectfont\raggedright
Each reader retains its FR-calibrated threshold for SP.
$\dagger$ GenPTW predicts no edited pixels on SP at its FR-calibrated threshold.
\par}
\end{table*}

As shown in Table~\ref{tab:thresholds}, the FR advantage remains clear when background false positives are controlled to comparable levels. ReliaGuard attains 85.61\% FR IoU at 1.07\% measured FPR, whereas APT$^*$ reaches 40.59\% at 1.02\% FPR. The paired gain is 45.02 percentage points (95\% CI: [43.32, 46.68]), accompanied by 92.21\% target recall.

Comparing fixed and calibrated operating points clarifies differences that SP results alone do not expose. At fixed thresholds, EditGuard and GenPTW achieve 81.78\% and 91.20\% SP IoU, but their FR background FPRs approach 100\%. Raising their thresholds to satisfy 1\% FR FPR suppresses much of their SP foreground response as well, reducing SP IoU to 0.62\% and 0.00\%. StableGuard-C retains the highest calibrated SP IoU (96.47\%) but reaches only 0.71\% on FR. For ReliaGuard, calibration changes FR IoU from 85.69\% to 85.61\% and SP IoU from 94.95\% to 94.90\%, indicating stable separation between target edits and reconstructed backgrounds.

\subsubsection{Robustness to Image Processing}
\begin{table*}[tp]
\centering
\caption{Edit localization and source-identifier recovery under ordinary propagation.}
\label{tab:prop}
\fontsize{8}{10}\selectfont
\setlength{\tabcolsep}{3.5pt}
\renewcommand{\arraystretch}{1.12}

\begin{tabular*}{\linewidth}{@{\extracolsep{\fill}}l*{10}{c}@{}}
\toprule
\multirow[c]{2}{*}[-0.6ex]{Method} &
\multicolumn{5}{c}{FR} &
\multicolumn{5}{c}{SP} \\
\cmidrule(lr){2-6}
\cmidrule(lr){7-11}
& IoU$\uparrow$ & F1$\uparrow$ & FPR$\downarrow$
& $\mathrm{BA}_{\mathrm{ID}}\uparrow$ & ER$\uparrow$
& IoU$\uparrow$ & F1$\uparrow$ & FPR$\downarrow$
& $\mathrm{BA}_{\mathrm{ID}}\uparrow$ & ER$\uparrow$ \\
\midrule
WAM & 2.04 & 3.59 & 1.44 & 90.28 & \underline{85.80} & 2.41 & 4.31 & 0.00 & \textbf{98.32} & \underline{99.98} \\
StableGuard-C & 1.88 & 3.25 & 5.15 & 50.26 & \textbf{100.00} & 37.78 & 42.07 & 13.52 & 70.56 & \textbf{100.00} \\
OmniGuard & 1.64 & 2.99 & 1.46 & \textnormal{--} & \textnormal{--} & 28.17 & 31.39 & 2.98 & \textnormal{--} & \textnormal{--} \\
EditGuard & 0.63 & 1.20 & 0.63 & 53.79 & \textbf{100.00} & 0.40 & 0.75 & 2.91 & 75.11 & \textbf{100.00} \\
GenPTW & 0.53 & 1.02 & 0.90 & \underline{91.78} & \textbf{100.00} & 0.00 & 0.00 & 0.00 & \underline{92.98} & \textbf{100.00} \\
APT & 19.90 & 29.54 & 5.05 & \textnormal{--} & \textnormal{--} & 19.46 & 28.52 & 3.17 & \textnormal{--} & \textnormal{--} \\
APT$^*$ & \underline{31.04} & \underline{40.42} & 0.90 & \textnormal{--} & \textnormal{--} & \underline{39.46} & \underline{49.39} & 0.75 & \textnormal{--} & \textnormal{--} \\
ReliaGuard & \textbf{81.01} & \textbf{88.45} & 1.76 & \textbf{92.76} & \textbf{100.00} & \textbf{86.49} & \textbf{91.51} & 0.80 & 92.79 & \textbf{100.00} \\
\bottomrule
\end{tabular*}

\par\smallskip
{\fontsize{7}{8.5}\selectfont\raggedright
ER denotes message extraction rate; $\mathrm{BA}_{\mathrm{ID}}$ is computed over extracted messages.
-- denotes no comparable source-message output.
\par}
\end{table*}

Across seven families of image processing, ReliaGuard retains 81.01\% FR IoU and 86.49\% SP IoU, compared with 31.04\% and 39.46\% for APT$^*$ (Table~\ref{tab:prop}). Identifier-bit accuracy remains 92.76\% for FR and 92.79\% for SP, with 100\% extraction rate in both conditions. WAM achieves higher SP identifier-bit accuracy (98.32\%), but its localization drops sharply at the transferred FR threshold, showing that message recovery and spatial localization respond differently to the same processing operations.

\begin{figure*}[tp]
\centering
\includegraphics[width=\textwidth]{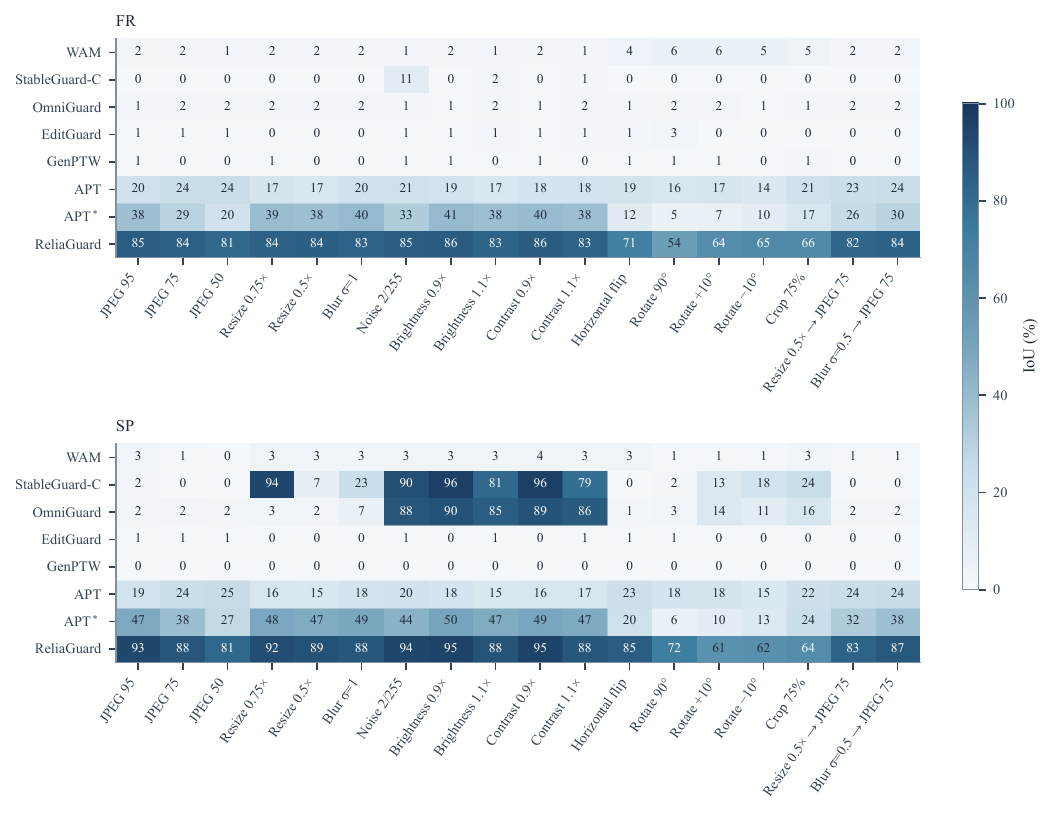}
\caption{FR and SP localization of active methods under 18 image-processing settings, using fixed FR-calibrated thresholds. Heatmap values are IoUs rounded to whole percentages.}
\label{fig:allprop}
\end{figure*}

Figure~\ref{fig:allprop} breaks down performance across the 18 settings. Compression and resizing have a smaller effect than large geometric transformations: JPEG compression at quality 75 gives 83.85\% FR and 88.03\% SP IoU, and half-resolution resizing gives 84.47\% and 89.47\%. By contrast, $90^\circ$ rotation lowers IoU to 53.66\% (FR) and 72.40\% (SP), while 75\% area crops yield 65.80\% FR and 64.24\% SP IoU, and 50\% area crops yield 48.77\% FR and 47.51\% SP IoU. On 120 sources evaluated under four ordered processing chains absent from training, ReliaGuard obtains 83.43\% FR and 86.40\% SP IoU.

\subsubsection{Protected-Image Quality}
\begin{table*}[tp]
\centering
\caption{Protected-image fidelity across three image domains.}
\label{tab:quality}
\fontsize{8}{10}\selectfont
\setlength{\tabcolsep}{3.5pt}
\renewcommand{\arraystretch}{1.12}

\begin{tabular*}{\linewidth}{@{\extracolsep{\fill}}l*{9}{c}@{}}
\toprule
\multirow[c]{2}{*}[-0.6ex]{Method} &
\multicolumn{3}{c}{COCO ($n=100$)} &
\multicolumn{3}{c}{Generated ($n=100$)} &
\multicolumn{3}{c}{ImageNet ($n=100$)} \\
\cmidrule(lr){2-4}
\cmidrule(lr){5-7}
\cmidrule(lr){8-10}
& PSNR (dB)$\uparrow$ & SSIM$\uparrow$ & LPIPS$\downarrow$
& PSNR (dB)$\uparrow$ & SSIM$\uparrow$ & LPIPS$\downarrow$
& PSNR (dB)$\uparrow$ & SSIM$\uparrow$ & LPIPS$\downarrow$ \\
\midrule
WAM & \underline{38.32} & \underline{0.9768} & \underline{0.0348} & \underline{37.80} & \underline{0.9796} & \underline{0.0238} & \underline{39.32} & \underline{0.9760} & \underline{0.0444} \\
StableGuard-C & 25.87 & 0.7328 & 0.0733 & 31.46 & 0.9429 & 0.0256 & 31.01 & 0.8580 & 0.0517 \\
OmniGuard & \textbf{46.27} & \textbf{0.9935} & \textbf{0.0005} & \textbf{47.52} & \textbf{0.9948} & \textbf{0.0003} & \textbf{49.53} & \textbf{0.9951} & \textbf{0.0005} \\
EditGuard & 31.91 & 0.8370 & 0.0474 & 31.36 & 0.8554 & 0.0414 & 31.99 & 0.8141 & 0.0794 \\
GenPTW & 25.72 & 0.7174 & 0.0803 & 30.21 & 0.9147 & 0.0354 & 30.64 & 0.8406 & 0.0567 \\
APT & 25.91 & 0.7121 & 0.3945 & 25.91 & 0.7641 & 0.3012 & 31.09 & 0.8392 & 0.2156 \\
ReliaGuard & 32.70 & 0.9066 & 0.1173 & 32.70 & 0.9207 & 0.0883 & 32.70 & 0.8867 & 0.2060 \\
\bottomrule
\end{tabular*}

\par\smallskip
{\fontsize{7}{8.5}\selectfont\raggedright
APT and APT$^*$ share a writer. Their scores include the native-resolution conversion.
\par}
\end{table*}

Across the 300 test sources, ReliaGuard obtains an average PSNR of 32.70~dB, SSIM of 0.9046, and LPIPS of 0.1372 (Table~\ref{tab:quality}). Because the quantization-aware projection enforces a per-image PSNR floor of 32~dB, PSNR remains consistent across domains, whereas LPIPS varies from 0.0883 on generated images to 0.2060 on ImageNet due to content-dependent perceptual sensitivity.

Image fidelity and reconstruction tolerance must be evaluated jointly. APT and APT$^*$, which share a writer, average 27.64~dB PSNR, 0.7718 SSIM, and 0.3038 LPIPS. Relative to APT, ReliaGuard improves PSNR by 5.06~dB and reduces LPIPS by 54.84\%, while also improving calibrated FR IoU over APT$^*$ by 45.02 percentage points. WAM (37.80--39.32~dB) and OmniGuard (46.27--49.53~dB) preserve higher image fidelity, but their calibrated FR IoUs drop to 1.50\% and 1.29\%. Stronger tolerance to full-output regeneration is thus accompanied by higher embedding distortion compared with the highest-fidelity baselines.

\begin{figure*}[tp]
\centering
\includegraphics[width=\textwidth]{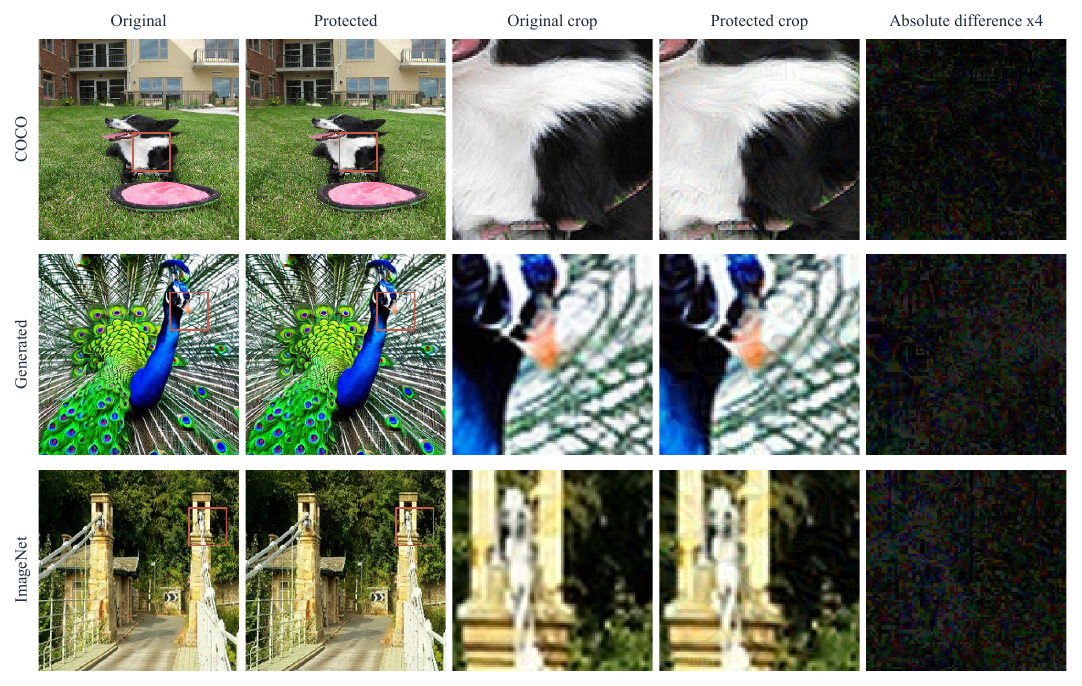}
\caption{Protected images, local crops, and embedding residuals from the three domains. Representative high-fidelity examples from each domain are shown. Original and protected images use the same crop; absolute differences are amplified fourfold. Table~\ref{tab:quality} reports quality over all 300 sources.}
\label{fig:quality_zoom}
\end{figure*}

Figure~\ref{fig:quality_zoom} shows protected images, magnified crops, and fourfold-amplified embedding residuals. Main scene structures are preserved at full-image scale, while the amplified difference maps visualize the spatial distribution of subtle texture perturbations.

\subsubsection{Effects of Target Size and Shape}

\begin{figure*}[tp]
\centering
\includegraphics[
    width=\textwidth,
    trim=0 2pt 0 2pt,
    clip
]{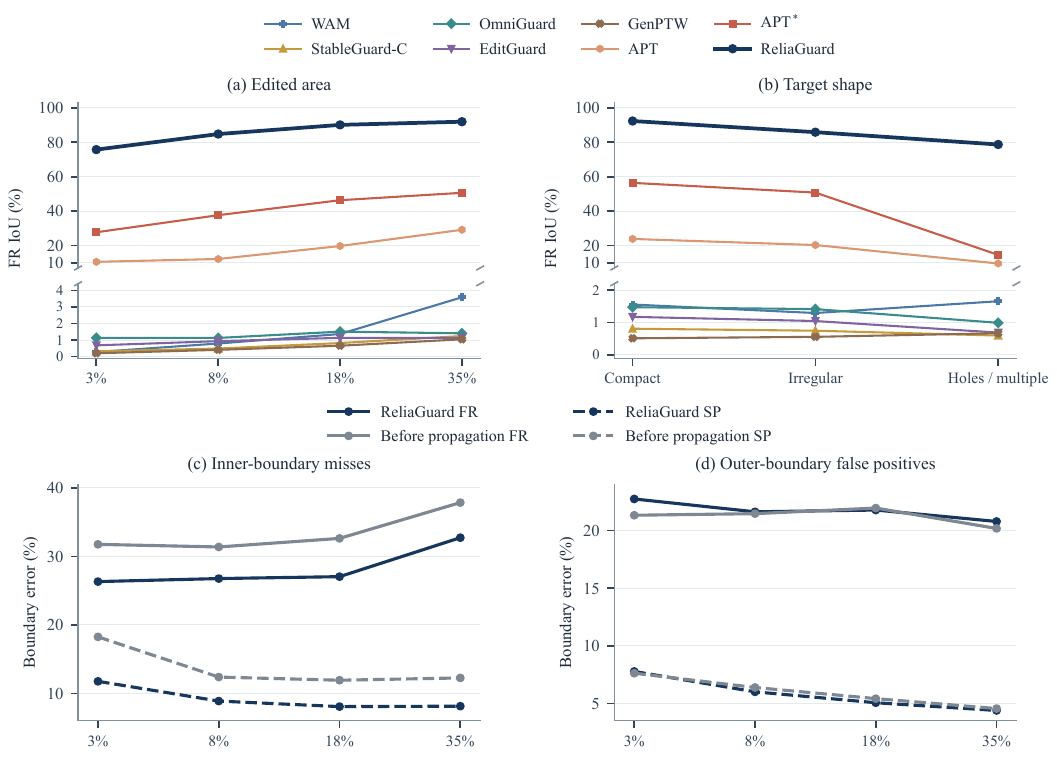}
\caption{Effects of target geometry. (a,b) FR IoU across target areas and shapes. (c,d) Inner-band FNR and outer-band FPR before and after propagation training, evaluated on unpropagated images. Solid and dashed curves denote FR and SP, respectively. Boundary bands are eight pixels wide.}
\label{fig:strata}
\label{fig:boundary}
\end{figure*}

Small and geometrically complex edits remain the most difficult cases (Fig.~\ref{fig:strata}). At 3\% target area, ReliaGuard reaches 75.71\% FR IoU compared with 27.73\% for APT$^*$. As target area grows to 8\%, 18\%, and 35\%, FR IoU increases to 84.74\%, 90.07\%, and 91.91\% (SP rises from 88.26\% to 98.40\%). Across compact, irregular connected, and hole/multiple-component masks, FR IoU is 92.31\%, 85.83\%, and 78.68\%, respectively. Averaged across geometries, inner-band FNR and outer-band FPR are 28.22\% and 21.74\% for FR, versus 9.21\% and 5.81\% for SP. Because boundary pixels occupy a larger fraction of small or multi-component masks, comparable contour displacements produce larger IoU reductions in these cases.

\subsubsection{Joint Source Attribution and Localization}
\begin{table}[tp]
\centering
\caption{Source attribution and joint localization performance.}
\label{tab:source}
\fontsize{8}{10}\selectfont
\setlength{\tabcolsep}{3.5pt}
\renewcommand{\arraystretch}{1.12}

\begin{tabular*}{\linewidth}{@{\extracolsep{\fill}}l*{4}{c}@{}}
\toprule
\multirow[c]{2}{*}[-0.6ex]{Condition} &
\multicolumn{2}{c}{Source recovery} &
\multicolumn{2}{c}{Joint success} \\
\cmidrule(lr){2-3}
\cmidrule(lr){4-5}
& $\mathrm{BA}_{\mathrm{code}}\uparrow$ & CA$\uparrow$
& @0.50$\uparrow$ & @0.75$\uparrow$ \\
\midrule

\multicolumn{5}{c}{\textbf{FR}} \\
Unpropagated & 90.58 & 99.67 & 96.33 & 90.78 \\
Propagated & 84.57 & 85.43 & 81.71 & 74.22 \\
\midrule

\multicolumn{5}{c}{\textbf{SP}} \\
Unpropagated & 100.00 & 100.00 & 99.67 & 95.67 \\
Propagated & 92.87 & 85.71 & 82.88 & 78.45 \\
\bottomrule
\end{tabular*}

\par\smallskip
{\fontsize{7}{8.5}\selectfont\raggedright
Scores are percentages. CA denotes correct identifier acceptance.
Joint@$t$ requires correct acceptance and per-case IoU $\ge t$.
Propagated results are averaged over the seven processing families.
\par}
\end{table}

Table~\ref{tab:source} reports joint attribution and localization performance. On unpropagated FR images, error correction raises bit accuracy from 90.58\% on coded bits to 100\% on the 16-bit identifier, with 99.67\% correct acceptance and 90.78\% joint success at $\mathrm{IoU}\ge 0.75$ (95.67\% for SP). After seven-family image processing, joint success at $\mathrm{IoU}\ge 0.75$ is 74.22\% for FR and 78.45\% for SP, with 85.43\% and 85.71\% correct acceptance.

\subsubsection{Transfer to Another Editing Protocol}
\begin{table*}[tp]
\centering
\caption{Localization under the APT SD1.5 protocol on 100 sources, using a fixed threshold of 0.5.}
\label{tab:transfer}
\fontsize{8}{10}\selectfont
\setlength{\tabcolsep}{3.5pt}
\renewcommand{\arraystretch}{1.12}

\begin{tabular*}{\linewidth}{@{\extracolsep{\fill}}l*{6}{c}@{}}
\toprule
\multirow[c]{2}{*}[-0.6ex]{Method} &
\multicolumn{3}{c}{FR} &
\multicolumn{3}{c}{SP} \\
\cmidrule(lr){2-4}
\cmidrule(lr){5-7}
& Edit IoU$\uparrow$ & Edit F1$\uparrow$ & Background IoU$\uparrow$
& Edit IoU$\uparrow$ & Edit F1$\uparrow$ & Background IoU$\uparrow$ \\
\midrule
APT & 67.27 & \underline{78.49} & 89.47 & 75.39 & 84.63 & 94.63 \\
APT$^*$ & \underline{67.42} & 77.86 & \underline{91.97} & \underline{78.45} & \underline{86.48} & \underline{94.88} \\
ReliaGuard & \textbf{81.73} & \textbf{89.04} & \textbf{96.22} & \textbf{95.38} & \textbf{97.36} & \textbf{99.08} \\
\midrule
Before propagation training & 83.11 & 90.15 & 96.59 & 95.44 & 97.30 & 99.12 \\
\bottomrule
\end{tabular*}

\par\smallskip
{\fontsize{7}{8.5}\selectfont\raggedright
Background scores use the authors' resized soft masks; edit scores use binary targets.
The before-propagation row is excluded from ranking.
\par}
\end{table*}

Without retraining, ReliaGuard achieves 81.73\% FR and 95.38\% SP edit-region IoU on the APT authors' 100-source Stable Diffusion 1.5 (SD1.5) protocol, outperforming APT$^*$ by 14.31 and 16.93 percentage points, respectively (Table~\ref{tab:transfer}), with background IoUs of 96.22\% (FR) and 99.08\% (SP). Reproduced APT and APT$^*$ background scores closely match their published results on the authors' $256\times256$ evaluation protocol.

Before propagation training, the same readout achieves 83.11\% FR and 95.44\% SP edit-region IoU on this protocol. Propagation training lowers clean SD1.5 FR IoU by 1.38 percentage points while improving seven-family processing means by 5.41 and 9.58 percentage points for FR and SP, respectively. The two comparisons reflect a trade-off between adaptation to distribution processing and transfer to an unseen clean editor, while the final reader still outperforms APT$^*$ by a wide margin.

\begin{table*}[tp]
\centering
\caption{Structural ablations trained from the same localization initialization.}
\label{tab:readout_structure}
\fontsize{8}{10}\selectfont
\setlength{\tabcolsep}{3.5pt}
\renewcommand{\arraystretch}{1.12}

\begin{tabular*}{\linewidth}{@{\extracolsep{\fill}}l*{8}{c}@{}}
\toprule
\multirow[c]{2}{*}[-0.6ex]{Variant} &
\multicolumn{5}{c}{FR} &
\multicolumn{3}{c}{SP} \\
\cmidrule(lr){2-6}
\cmidrule(lr){7-9}
& IoU$\uparrow$ & F1$\uparrow$ & FPR$\downarrow$
& Inner FNR$\downarrow$ & Outer FPR$\downarrow$
& IoU$\uparrow$ & F1$\uparrow$ & FPR$\downarrow$ \\
\midrule
Full architecture & \underline{83.17} & \underline{90.16} & 1.15 & 31.93 & 22.26 & \underline{93.33} & \underline{96.30} & 0.28 \\
Without BER & 78.91 & 87.30 & 1.11 & 41.84 & 20.82 & 81.90 & 89.16 & 0.37 \\
Without background contrast & 83.08 & 90.08 & 1.15 & 31.97 & 22.42 & 93.32 & 96.29 & 0.29 \\
Without NSR & 82.31 & 89.62 & 1.18 & 33.57 & 22.84 & 91.50 & 95.21 & 0.37 \\
Bilinear reassembly & \textbf{83.18} & \textbf{90.16} & 1.15 & 31.88 & 22.28 & \textbf{93.38} & \textbf{96.33} & 0.28 \\
\bottomrule
\end{tabular*}

\par\smallskip
{\fontsize{7}{8.5}\selectfont\raggedright
Scores are percentages. Inner FNR and outer FPR refer to the eight-pixel boundary bands.
Bold and underlined values denote the best and second-best overlap scores.
All variants use the same initialization, data, and 100-epoch training budget, with evaluation on 144 sources and threshold calibration on 72 separate sources.
Paired IoU differences and confidence intervals are shown in Fig.~\ref{fig:origin_mechanisms}.
\par}
\end{table*}

\subsection{Ablation Studies}

\subsubsection{Spatial Evidence and Key Correspondence}

\begin{figure*}[tp]
\centering
\includegraphics[width=\textwidth]{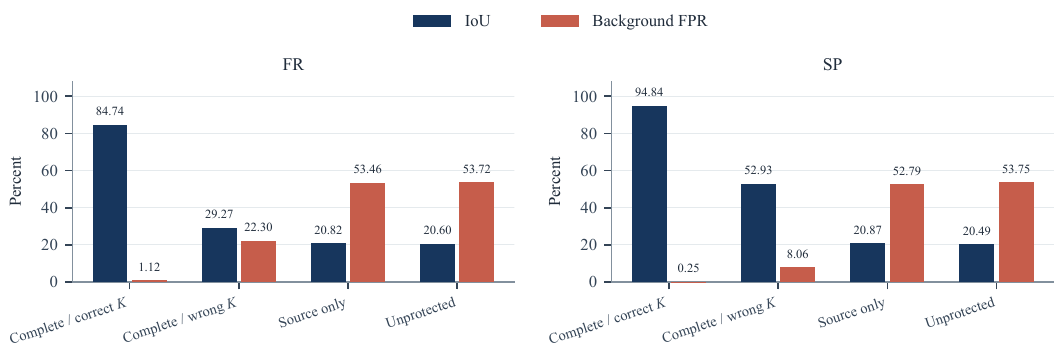}
\caption{Effect of spatial evidence and key correspondence on 60 sources. Reader weights and the decision threshold are fixed. Wrong-key results are averaged over five alternative keys.}
\label{fig:evidence_controls}
\end{figure*}

Recovering a source message alone is insufficient for blind edit localization (Fig.~\ref{fig:evidence_controls}). Complete protection with the correct key achieves 84.74\% FR IoU at 1.12\% FPR. Removing spatial evidence while keeping the source watermark reduces IoU to 20.82\% and raises FPR to 53.46\% even though identifier-bit accuracy remains 100\%, close to the 20.60\% IoU obtained without protection. Reading protected images with alternative keys reduces IoU to 29.27\% (22.30\% FPR). Both the written spatial evidence and its correspondence with the reading key are therefore required for localization.

\subsubsection{Architectural Attribution within the Unified Reader}

\begin{figure}[tp]
\centering
\includegraphics[width=\linewidth]{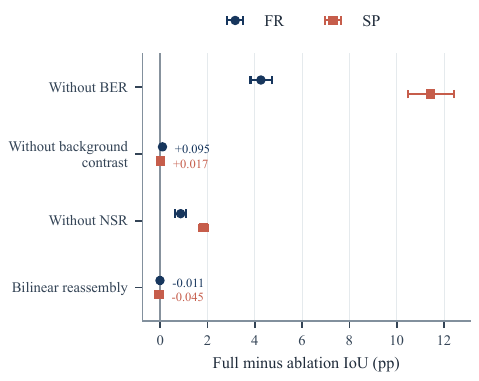}
\caption{Paired IoU differences between the full architecture and structural ablations. Positive values favor the full architecture. Error bars are 95\% confidence intervals from 5,000 source-group bootstrap resamples on the 144-source ablation set.}
\label{fig:origin_mechanisms}
\end{figure}

To evaluate the contribution of each readout stage and its internal operators, Table~\ref{tab:readout_structure} and Fig.~\ref{fig:origin_mechanisms} compare five architectural variants trained from matched initializations with the same data and 100-epoch training budget.

Removing the contextual residual refinement stage (\emph{Without BER}) produces the largest drop, reducing FR IoU from 83.17\% to 78.91\% ($-4.26$~pp) and SP IoU from 93.33\% to 81.90\% ($-11.43$~pp), while raising FR inner-boundary FNR from 31.93\% to 41.84\%. When only the background-weighting term is removed while retaining the encoder--decoder and feature inputs (\emph{Without background contrast}), IoU remains at 83.08\% for FR and 93.32\% for SP. Most of the gain in this stage is therefore attributable to the complete contextual residual path rather than the isolated weighting term.

Removing the multiscale readout stage (\emph{Without NSR}) reduces FR IoU by 0.86 percentage points (to 82.31\%) and SP IoU by 1.83 percentage points (to 91.50\%). Replacing learned reassembly inside NSR with standard bilinear interpolation (\emph{Bilinear reassembly}) yields nearly identical scores (83.18\% FR IoU and 93.38\% SP IoU). Taken together, these controlled ablations indicate that ReliaGuard's gains arise primarily from the two complete readout stages, contextual residual refinement and native-guided multiscale fusion, and that lightweight bilinear upsampling can substitute for learned reassembly in latency-sensitive deployments without loss of accuracy.

\subsubsection{Paired Propagation Training}

Compared with the reader prior to propagation training, paired propagation training improves the seven-family mean IoU by 5.41 percentage points for FR and 9.58 percentage points for SP. FR recall rises from 86.01\% to 90.05\% while FPR drops from 2.76\% to 1.76\%, and SP recall rises from 81.28\% to 91.18\% while FPR drops from 1.28\% to 0.80\%. On unpropagated images, inner-boundary FNR in Fig.~\ref{fig:boundary} also improves across all four target areas. Under matched initialization and training budgets on 96 sources across seven representative processing conditions, omitting propagated views lowers FR IoU from 80.68\% to 77.23\% and SP IoU from 85.09\% to 78.48\%, whereas the preservation loss and progressive augmentation schedule have smaller effects.
\section{Conclusion}

In this paper, we presented ReliaGuard, a reconstruction-aware watermarking framework designed to resolve background reconstruction ambiguity in joint source attribution and blind edit localization. By coordinating dual-view keyed evidence writing across protected and VAE-reconstructed images with a unified coarse-to-fine blind reader that combines contextual residual refinement and native-guided multiscale decoding, ReliaGuard preserves a frozen source-attribution channel while reliably separating intended edits from reconstructed backgrounds without requiring a reference image at inference. Extensive evaluations across three image domains, paired full-output and locally spliced editing regimes, seven image-processing families, and cross-protocol transfer demonstrate substantial gains in full-output localization accuracy alongside reliable identifier recovery, while controlled architectural ablations confirm that the primary improvements stem from the complete contextual and multiscale readout stages and show that lightweight bilinear upsampling can replace learned reassembly without sacrificing accuracy.

Despite these advances, localizing very small or intricate regions under severe geometric transformations remains challenging, and strong reconstruction tolerance still entails higher embedding distortion than the highest-fidelity baselines. Future work will focus on reducing embedding distortion, accelerating write-side optimization, and strengthening geometric invariance for practical deployment.

\end{document}